\pdfoutput=1
\documentclass[runningheads,a4paper]{llncs}

\usepackage{amssymb}
\usepackage{graphicx}
\usepackage{amsmath}
\usepackage{xcolor}
\usepackage{subfigure}
\usepackage{wrapfig}

\usepackage{url}
\urldef{\mailsa}\path|{alfred.hofmann, ursula.barth, ingrid.haas, frank.holzwarth,|
\urldef{\mailsb}\path|anna.kramer, leonie.kunz, christine.reiss, nicole.sator,|
\urldef{\mailsc}\path|erika.siebert-cole, peter.strasser, lncs}@springer.com|

\newcommand{\RN}[1]{%
	\textup{\lowercase\expandafter{\it \romannumeral#1}}%
}

\begin{document}

\mainmatter  

\title{Dynamic Routing on Deep Neural Network for Thoracic Disease Classification and Sensitive Area Localization}

\titlerunning{Dynamic routing on ChestX-ray Classification}

%
%
\author{Yan Shen, Mingchen Gao}
\authorrunning{}

\institute{Department of Computer Science and Engineering, \\
University at Buffalo
}

%
%

\toctitle{Dynamic routing on ChestX-ray Classification}
\tocauthor{}
\maketitle

\begin{abstract}
We present and evaluate a new deep neural network architecture for automatic thoracic disease detection on chest X-rays. Deep neural networks have shown great success in a plethora of visual recognition tasks such as image classification and object detection by stacking multiple layers of convolutional neural networks (CNN) in a feed-forward manner. However, the performance gain by going deeper has reached bottlenecks as a result of the trade-off between model complexity and discrimination power. We address this problem by utilizing the recently developed routing-by agreement mechanism in our architecture. A novel characteristic of our network structure is that it extends routing to two types of layer connections (1) connection between feature maps in dense layers, (2) connection between primary capsules and prediction capsules in final classification layer. We show that our networks achieve comparable results with much fewer layers in the measurement of AUC score. We further show the combined benefits of model interpretability by generating Gradient-weighted Class Activation Mapping (Grad-CAM) for localization. We demonstrate our results on the NIH chestX-ray14 dataset that consists of 112,120 images on 30,805 unique patients including 14 kinds of lung diseases.
\end{abstract}

\section{Introduction}

It is a relatively easy task for radiologists to read and diagnose chest X-ray images. However, teaching a computer to process hospital-scale of chest X-ray scans is extremely challenging. Chest X-rays is the most common imaging examinations in practice, with approximately 2 billion procedures per year \cite{rajpurkar2017chexnet}. The success of chest X-ray disease detection will lay the groundwork for more complex systems to provide consistent, trustable and interpretable second opinions on reading medical images of all kinds of modalities. 

Deep Learning methods have been applied to disease classification, sensitive area localization and tissue segmentation \cite{litjens2017survey}. The success of deep learning has made computer program an indispensable aid to physicians for disease analysis \cite{shin2016deep}. ``ChestX-ray14'' is so far the largest publicly available chest X-rays dataset \cite{wang2017chestx}. Along with the collection of the dataset, baseline models were also tested on this dataset. The best is a 50 layers ResNet. There are many followed up works on this dataset, such as DenseNet based models \cite{rajpurkar2017chexnet,yao2017learning} or attention guided CNN to integrate disease-specific region and global cues \cite{guan2018diagnose}. However most of the current work randomly split the data into training, validation and testing. It is likely to have images from the same patient appear in both training and testing set. Such experimental setting makes the direct comparisons of reported evaluation metrics problematic. Yao \cite{yao2018weakly} uses a learnable Log-Sum-Exp pooling functions in their network for classification and use Log-Sum-Exp pooling function to generate salient maps at different resolutions to indicate regions of interest (ROI). We also follow the split suggestion by Wang \cite{wang2017chestx,yao2018weakly} and does not use additional training data. 

A highly correlated task with disease classification is to localize the sensitive area related to diseases. Weakly-supervised pathology localization has been used to generate heatmap based on class activation mappings (CAMs) \cite{zhou2016learning}. RR Selvaraju \cite{selvaraju2017grad} used Gradient-weighted Class Activation Mapping (Grad-CAM) as a more generalized form of CAMs without the need of global average pooling at last layer of feature maps. Zhe et al. proposed a unified approach to simultaneously perform disease identification and localization \cite{li2017thoracic}. 

Current advancement in deep network's recognition power is typically achieved by going deeper with more layers and denser connections. One exception is Hinton's Capsule net \cite{sabour2017dynamic} which shows promising potential by its novel structure. The network's connectivity adapts to the coherence of input feature vectors other than being optimized through back propagation. The activations in higher levels are achieved by routing-by-agreement iteration. Dilin et al. \cite{wang2018optimization} views Capsule net as minimizing a clustering loss function with a KL divergence regularization iteratively. Capsule network has been extended to many applications, includes but not limited to pathology lung segmentation \cite{lalonde2018capsules} and brain tumor type classification \cite{afshar2018brain}.

Inspired by Hinton's work, this paper proposes a new implementation of Capsule net on CNNs. Our model involves three key contributions. 
\vspace{-0.2cm}
\begin{itemize}
	\item We introduce dense connectivities with dynamic routing into our network. Dense connectivity is achieved by a $1\times1$ convolutional layer that takes all of the previous feature maps as input. And we extend the routing-by-agreement mechanism to that $1\times1$ convolutional layer. This preserves DenseNet's nice property of facilitating training process while incorporates Capsule Net's routing mechanism to select more relevant feature maps in a bottom-up fashion. To the best of our knowledge, our paper is the first work that extends dynamic routing to convolutional layers.
	\item Our model is efficiently implemented using kernel trick. Feature maps need only to be calculated once per layer. The routing coefficient is set to be trainable only at the last iteration. Such implementation reduces the time for training and inference as its complexity is comparable to a single layer without routing iterations.
	\item Rather than generating heatmap before global average pooling layers or fully connected layer, we generate heatmap before an average pooling layer of strides $4\times4$ and a fully connected dynamic routing layer before prediction layer. Our generated heatmap preserves the benefit of model interpretability as CAM without sacrifice classification accuracy by introducing global average pooling.
\end{itemize}

\section{Methods}

In our networks, chest X-ray images are firstly pre-processed and then passed to a down-sampling block of Conv-Pool-Conv-Pool. The first convolutional layer is of size 7 and stride 2. And then the second pooling layer is using max-pooling of size 3 and stride 2. Following with the max-pooling layer, we use a convolutional layer of size 1 and stride 2. Finally we use average-pooling layer of size 2 and stride 2 before feeding to our dense layer. 
\begin{wrapfigure}{R}{7cm}
	\centering
	\includegraphics[width=\linewidth]{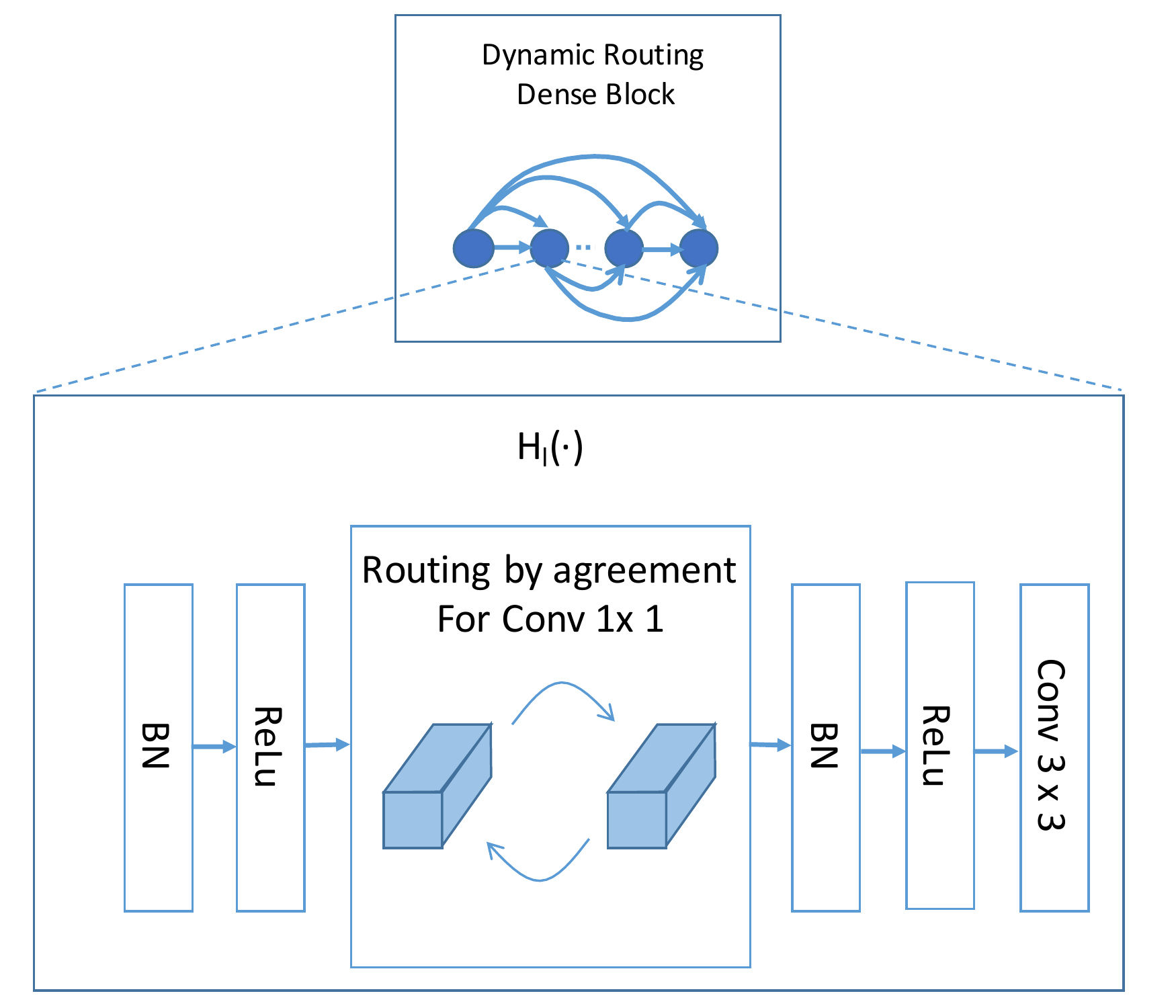}
	\caption{Each dense block consists 8 layers of composite functions. 1$\times$1 convolutional layer is updated using dynamic routing.} \label{fig:Conv1x1Routing}
\end{wrapfigure}

We use dense blocks after down-sampling blocks. Our dense block follows the pattern in \cite{huang2017densely} except for the 1$\times$1 convolutional layer. A dense layer consists of consecutive layers of composite functions which takes concatenated output produced in previous layers. Each composite function $H_l(\cdot)$ consists 6 consecutive operations: BN-ReLu-Conv(1$\times$1)-BN-ReLu-Conv(3$\times$3). In our network, dynamic routing is included between the connections of 1$\times$1 convolution layer. A dynamic routing dense block is illustrated in Fig. \ref{fig:Conv1x1Routing}.

Subsequent to dense blocks, we use a larger size convolutional layer with size 9 and stride 1. Then we use a average pooling layer of size 4 and stride 4 before we use a fully connected capsule layer to class labels. In our fully connected capsule layer, we reshape the feature map to primary capsules by taking 8 consecutive feature maps of each pixel as one capsule. Then we route the fully connected layer between primary capsules and disease label capsule following the routing by agreement mechanism in \cite{sabour2017dynamic}. Finally, we take the $L_2$ norm of each vectors in digit capsules as the digit of each disease label.

\subsection{ $1 \times 1$ Convolutional Capsule Layer}
In Capsule net, the coupling coefficient is updated iteratively as the agreements between input and output from the layer below. We extend the routing-by-agreement mechanism \cite{sabour2017dynamic} to $1\times 1$ convolutional layer. The output capsule vector and routing coefficient can be computed element-wisely. However this brute-force effort is computational exhaustive and not feasible. The convolutional kernels and coupling coefficient share the property of parameter sharing. We propose to use kernel trick to implement dynamic routing on feature maps efficiently. Recall that in $1\times 1$ convolutional layer, every output of feature map $\mathbf{g}_j$ is the linear combination of input feature maps $\mathbf{f}_i$.
\begin{equation}
\mathbf{g}_{j}=\sum_{i}K_{i,j}\mathbf{f}_{i},
\end{equation}
where $K_{i,j}$ is the scalar element of convolutional kernel. Here we follow the dynamic routing and define the term $\hat{\mathbf{f}}_{j|i}=W_{i,j}\mathbf{f}_i$ as the ``prediction vector" from input feature map $\mathbf{f}_i$ to output feature map $\mathbf{g}_j$. Similarly as the mechanism of capsule net, we use coupling coefficient to represent the agreements between the input and output feature map of $1\times 1$ convolutional layer. Specifically, the output feature maps is weighted sum of prediction vectors weighted on coupling coefficient.
\begin{equation}
\mathbf{g}_j=\sum_{i}c_{i,j}\hat{\mathbf{f}}_{j|i}
=\sum_{i}W_{i,j}c_{i,j}\mathbf{f}_i,
\end{equation}
where the coupling coefficient term $c_{i,j}$ is updated by the following two steps:
\textbf{Softmax Step}:
\begin{equation}
c_{i,j}= \frac{\mathrm{exp}(b_{i,j})}{\sum_{k}\mathrm{exp}(b_{i,k})}
\end{equation}
\textbf{Evidence Update Step}:
\begin{equation}
b_{i,j}\leftarrow b_{i,j}+\hat{\mathbf{f}}_{j|i}\cdot \mathrm{squash} (\mathbf{g}_j) 
\end{equation}

Rather than getting the new couping coefficient by updating the whole feature map in every iteration. We can take the simplified step by applying kernel tricks.
\begin{equation}
\hat{\mathbf{f}}_{j|i}\cdot \mathrm{squash} (\mathbf{g}_j)  = \frac{|\mathbf{g}_j|}{1+|\mathbf{g}_j|^2} \hat{\mathbf{f}}_{j|i} \cdot \mathbf{g}_j
\end{equation}

The term $\hat{\mathbf{f}}_{j|i} \cdot \mathbf{g}_j$ can be computed as:
\begin{align*}
\hat{\mathbf{f}}_{j|i} \cdot \mathbf{g}_j &= 
\sum_{l} \hat{\mathbf{f}}_{j|i} \cdot c_{i,j}\hat{\mathbf{f}}_{j|l} \\
&= \sum_{l} W_{i,j}W_{l,j}c_{l,j}\mathbf{f}_l \cdot \mathbf{f}_i 	
\end{align*}

And the norm of feature maps $|\mathbf{g}_j|$ can be computed as the weighted sum of $\hat{\mathbf{f}}_{j|i}\cdot \mathbf{g}_j$ :
\begin{equation}
|\mathbf{g}_j|^2 = \mathbf{g}_j\cdot\mathbf{g}_j
=  \sum_{i=1} c_{i,j}\mathbf{g}_j\cdot\hat{\mathbf{f}}_{j|i} 
\end{equation}

So we only need vector product of input feature maps $\mathbf{f}_l \cdot \mathbf{f}_i$, convolutional kernel $W_{i,j}$ and routing  coefficient $\mathbf{c}_{i,j}$ produced in last step to compute the term $\hat{\mathbf{f}}_{j|i}\cdot \mathrm{squash} (\mathbf{g}_j)$ to update routing coefficient. The inner product of input feature maps $\mathbf{f}_i\cdot\mathbf{f}_j$ only need to be computed once and are shared in every step of iteration.

\section{Experiment Results}
ChestX-ray14 dataset includes front view of chest X-ray images. Each one is annotated with multiple of 14 categories of lung diseases. We augment our training data by flipping the training images, randomly adjusting their brightness and contrast. To make our training and inferencing tractable, we resize the original chest X-ray image from original resolution $1024\times 1024$ to $256\times 256$. Those images are then standardized to zero mean and unit scale as the first step of our network. To validate the performance of our model, we follow the training and testing partition suggested by \cite{wang2017chestx} with 86524 training/validation images and 25596 testing images.
\begin{table}
	\centering
	\setlength{\tabcolsep}{3pt}
	\hspace*{-0.35cm}
	\setlength{\marginparwidth}{25pt}
	\begin{tabular}{ | c | c | c | c | c | c}
		\hline
		\textbf{Pathology}  & \textbf{Wang et al.\cite{wang2017chestx}} &\textbf{Yao et al. \cite{yao2018weakly}} & \textbf{Our Proposed} & \textbf{Our Baseline}\\ \hline
		Atelectasis & 0.7003  &  0.733 & \textbf{0.766} & 0.616\\
		Cardiomegaly & 0.8100  &  \textbf{0.856} & 0.801 & 0.761\\
		Effusion & 0.7585 &  \textbf{0.806} & 0.797 & 0.710\\  
		Infiltration & 0.6614 & 0.673  & \textbf{0.751} & 0.611\\
		Mass & 0.6933 & \textbf{0.777} &  0.760 & 0.589\\
		Nodule & 0.6687 & 0.718 & \textbf{0.741} & 0.534\\
		Pneumonia & 0.6580 &0.684  &  \textbf{0.778} & 0.569\\
		Pneumothorax & 0.7993 & \textbf{0.805} & 0.800 & 0.662\\
		Consolidation & 0.7032  &0.711 & \textbf{0.787} & 0.617\\
		Edema & 0.8052 &0.806   & \textbf{0.820} & 0.744\\
		Emphysema & 0.8330 & \textbf{0.842}  & 0.773 & 0.672\\
		Fibrosis & \textbf{0.7859} & 0.743& 0.765 & 0.630\\
		Pleural Thickening & 0.6835 &0.724  & \textbf{0.759} & 0.611\\
		Hernia & \textbf{0.8717} &0.775 & 0.748 & 0.441\\ \hline
		Average & 0.738 & 0.761  & \textbf{0.775} & 0.626 \\
		\hline
	\end{tabular}
	\vspace{0.2cm}
	\caption{Compared to our baseline model, our proposed model achieves performance increase with replacing the $1\times 1$ convolutional layer with our $1\times 1$ capsule convolutional layer. The gain is brought by the increased generalization ability through routing-by-agreement between capsules. Our results also outperforms  state-of-the-art algorithm in the literature.  }
	\label{table:aucscore}
\end{table}

Our neural network model is implemented using Tensorflow. Models are trained using Adam optimizer. The learning rate is set to  $\alpha=0.001$, $\beta_1=0.9$, $\beta_2=0.999$ and $\epsilon=10^{-8}$ as our default parameters. Parameters are initialized using random normal initializations.

Curriculum learning is used in our training to stabilize our training process \cite{bengio2009curriculum}. As sparse positive labels in training data favors negative prediction, we set a down-scaling parameter on negative labels to compensate for that. We firstly set $\lambda_{+}=1$ and $\lambda_{-}=0.05$. And then we shifted to $\lambda_{+}=\frac{|N|}{|P|+|N|}$ and $\lambda_{-}=\frac{|P|}{|P|+|N|}$ after 50 epochs. Our whole training takes around $400,000$ global steps from random initialization and converges in $150,000$ global steps. We use GTX-1080Ti to accelerate our training process. It takes about 30 hours for our whole training process. We train our network only on chest X-ray dataset without any pre-training. 

We explore the impact of dynamic routing on 1$\times$1 convolutional capsules and the variations of network architectures on the performance of disease classifications. We replace the 1$\times$1 convolutional capsule in our proposed model to standard 1$\times$1 convolutional in our baseline. All the variations in network architectures are trained using the same settings to compare apple to apple. 

Finally, we generate Grad-CAM for as interpretation of our model's prediction. We investigate these regions that are considered important for our disease predictions and compare it with the bounding box that are provided by professional physicians. To define the important region of our generated Grad-CAM, we normalize our Grad-CAM from 0 to 1, and preserve those areas with an activation larger than 0.1 as the important region. 



\textbf{Classification Accuracy}
 Many of the follow up works on this dataset split the dataset randomly rather follow the suggestion given in the original dataset \cite{wang2017chestx}. We compare our models with the only two utilizing the official splits of training and testing data. Our model is demonstrated to be the state-of-the-art as shown in Table \ref{table:aucscore}.




Convolutional capsule net achieves stable results in every category of pathology label predictions with much smaller number of layers and simpler network structures. Replacing our 1$\times$1 convolutional capsules with standard 1$\times$1 convolutional layer results in degraded accuracies in nearly all categories of pathology. In average, the standard standard 1$\times$1 convolutional layer reduced accuracy by $15\%$. The increased accuracy reported in test dataset by our proposed model demonstrated the effectiveness of adopting capsule routing in convolutional layers. 

\begin{figure}[hbtp]
	\centering
	\subfigure[]
	  {\includegraphics[width=0.35\textwidth]{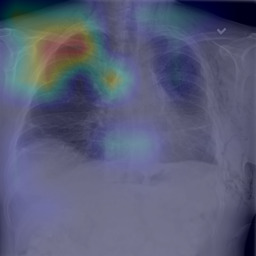}
	  	\label{fig:grad_cam11}
	  }
	\subfigure[]
	  {\includegraphics[width=0.35\textwidth]{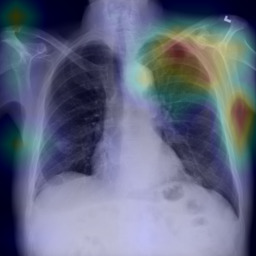}
	  	\label{fig:grad_cam12}
	  	}\\
	\caption{Two Patients with lung Atelectasis. The dynamic routing dense model along with Grad-CAM identifies the left or right upper lung Atelectasis, respectively and correctly classifies the pathology.}
	\label{fig:grad_cam1} 
\end{figure}

\textbf{Disease Localization}
We generate heap-map to visualize the area that is indicative of a suspect disease. We use the Grad-CAM to generate heat-map for disease area localization. We generate Grad-CAM $M_c$ from primary capsules at the resolution of $32\times32$ before $4\times 4$ average pooling layer. And then we up-sample it to the dimensions of input image and overlay it with the corresponding images. Fig. \ref{fig:grad_cam1} shows the heat-map generated by two chest x-ray images of patients diagnosed with Atelectasis. Qualitatively, the Grad-CAM of our model almost overlaps with the sensitive area of lung that are diagnosed with the corresponding pathology. Specifically in \ref{fig:grad_cam11}, the patient has Atelectasis in upper part of his left lung. The Grad-CAM of his chest X-ray image overlaps with his upper-left lung. Similarly in \ref{fig:grad_cam12} generated Grad-CAM for patient with upper-right lung pathology is activated at upper-right lung. We find that the heat-map generated at primary capsule level is indicative to disease area even though it is generated at low resolution.

For a quantitative analysis, We compare our generated Grad-CAM with the hand annotated ground truth (GT) boxes included in ChestX-ray14. Although the total number of B-Box annotations (1600 images) is relatively small compared with the entire dataset, it is still reasonable to  estimate on the interpretation of our model. To exam the accuracy of our computed Grad-CAM versus the GT B-Box, we use Intersection over the detected B-Box ratio (IoBB) for measurement. Table \ref{table:localization} illustrates the localization accuracy (Acc.) for each disease type, with $T(IoBB) \in \{0.1, 0.25, 0.5\}.$

\begin{table}
	\centering
	\setlength{\tabcolsep}{3pt}
	\hspace*{-3.2cm}	\begin{tabular}{ | c | c | c | c | c | c | c | c | c |}
		\hline
		\textbf{T(Iobb)}  & \textbf{Atelectasis} & \textbf{Cardiomegaly}  & \textbf{Effusion} & \textbf{Infiltration} & \textbf{Mass}& \textbf{Nodule}& \textbf{Pneumonia}& \textbf{Pneumothorax}\\ \hline \hline
		\multicolumn{9}{|c|}{T(IoBB)=0.1} \\
		\hline
		\textbf{Acc.} & 0.6977 & 0.8333 & 0.6234 & 0.635 & 0.4324 & 0.1234 & 0.6973 & 0.4687 \\
		\hline
		\multicolumn{9}{|c|}{T(IoBB)=0.25} \\
		\hline
		\textbf{Acc.} & 0.4534 & 0.8277 & 0.4840 & 0.5734 & 0.3866 & 0.0023 & 0.5342 & 0.3512 \\
		\hline
		\multicolumn{9}{|c|}{T(IoBB)=0.5} \\
		\hline
		\textbf{Acc.} & 0.2198 & 0.5231 & 0.2473 & 0.2412 & 0.1854 & 0.0019 & 0.3693 & 0.0716 \\
		\hline
	\end{tabular}

\hspace*{-3.2cm}
	\caption{Pathology localization accuracy for 8 disease classes. Because our primary capsule only have a resolution of $8\times 8$, we use the layer before $4\times 4$ average pooling that have a resolution of $32 \times 32$. Our generated Grad-CAM is like neuralization on CAM and Grad-CAM that trades-off on model interpretation and classification}
	\label{table:localization}
\end{table}

\vspace{-3em}
\section{Conclusion}
In this work, we handle the disease detection problem by using dynamic routing between $1\times1$ convolutional layers in dense block. We further test our network's detection accuracy and model interpretability in our experiment. For future work, we plan to improve disease localization by integrating location information provided in the dataset using semi-supervised learning. 

\bibliographystyle{abbrv}
\bibliography{MLMI2018_Capsule_XRAY}

\begin{thebibliography}{10}

\bibitem{afshar2018brain}
P.~Afshar, A.~Mohammadi, and K.~N. Plataniotis.
\newblock Brain tumor type classification via capsule networks.
\newblock {\em arXiv preprint arXiv:1802.10200}, 2018.

\bibitem{bengio2009curriculum}
Y.~Bengio, J.~Louradour, R.~Collobert, and J.~Weston.
\newblock Curriculum learning.
\newblock In {\em ICML}, pages 41--48. ACM, 2009.

\bibitem{guan2018diagnose}
Q.~Guan, Y.~Huang, Z.~Zhong, Z.~Zheng, L.~Zheng, and Y.~Yang.
\newblock Diagnose like a radiologist: Attention guided convolutional neural
  network for thorax disease classification.
\newblock {\em arXiv preprint arXiv:1801.09927}, 2018.

\bibitem{huang2017densely}
G.~Huang, Z.~Liu, L.~van~der Maaten, and K.~Q. Weinberger.
\newblock Densely connected convolutional networks.
\newblock In {\em CVPR}, 2017.

\bibitem{lalonde2018capsules}
R.~LaLonde and U.~Bagci.
\newblock Capsules for object segmentation.
\newblock {\em arXiv preprint arXiv:1804.04241}, 2018.

\bibitem{li2017thoracic}
Z.~Li, C.~Wang, M.~Han, Y.~Xue, W.~Wei, L.-J. Li, and F.-F. Li.
\newblock Thoracic disease identification and localization with limited
  supervision.
\newblock {\em CVPR}, 2017.

\bibitem{litjens2017survey}
G.~Litjens, T.~Kooi, B.~E. Bejnordi, A.~A.~A. Setio, F.~Ciompi, M.~Ghafoorian,
  J.~A. van~der Laak, B.~van Ginneken, and C.~I. S{\'a}nchez.
\newblock A survey on deep learning in medical image analysis.
\newblock {\em Medical image analysis}, 42:60--88, 2017.

\bibitem{rajpurkar2017chexnet}
P.~Rajpurkar, J.~Irvin, K.~Zhu, B.~Yang, H.~Mehta, T.~Duan, D.~Ding, A.~Bagul,
  C.~Langlotz, K.~Shpanskaya, et~al.
\newblock Chexnet: Radiologist-level pneumonia detection on chest x-rays with
  deep learning.
\newblock {\em arXiv preprint arXiv:1711.05225}, 2017.

\bibitem{sabour2017dynamic}
S.~Sabour, N.~Frosst, and G.~E. Hinton.
\newblock Dynamic routing between capsules.
\newblock In {\em NIPS}, pages 3859--3869, 2017.

\bibitem{selvaraju2017grad}
R.~R. Selvaraju, M.~Cogswell, A.~Das, R.~Vedantam, D.~Parikh, and D.~Batra.
\newblock Grad-cam: Visual explanations from deep networks via gradient-based
  localization.
\newblock In {\em CVPR}, pages 618--626, 2017.

\bibitem{shin2016deep}
H.-C. Shin, H.~R. Roth, M.~Gao, L.~Lu, Z.~Xu, I.~Nogues, J.~Yao, D.~Mollura,
  and R.~M. Summers.
\newblock Deep convolutional neural networks for computer-aided detection: Cnn
  architectures, dataset characteristics and transfer learning.
\newblock {\em TMI}, 35(5):1285--1298, 2016.

\bibitem{wang2018optimization}
D.~Wang and Q.~Liu.
\newblock An optimization view on dynamic routing between capsules.
\newblock {\em ICLR workshop 2018}.

\bibitem{wang2017chestx}
X.~Wang, Y.~Peng, L.~Lu, Z.~Lu, M.~Bagheri, and R.~M. Summers.
\newblock Chestx-ray8: Hospital-scale chest x-ray database and benchmarks on
  weakly-supervised classification and localization of common thorax diseases.
\newblock In {\em CVPR}, pages 3462--3471. IEEE, 2017.

\bibitem{yao2017learning}
L.~Yao, E.~Poblenz, D.~Dagunts, B.~Covington, D.~Bernard, and K.~Lyman.
\newblock Learning to diagnose from scratch by exploiting dependencies among
  labels.
\newblock {\em arXiv preprint arXiv:1710.10501}, 2017.

\bibitem{yao2018weakly}
L.~Yao, J.~Prosky, E.~Poblenz, B.~Covington, and K.~Lyman.
\newblock Weakly supervised medical diagnosis and localization from multiple
  resolutions.
\newblock {\em arXiv preprint arXiv:1803.07703}, 2018.

\bibitem{zhou2016learning}
B.~Zhou, A.~Khosla, A.~Lapedriza, A.~Oliva, and A.~Torralba.
\newblock Learning deep features for discriminative localization.
\newblock In {\em CVPR}, pages 2921--2929. IEEE, 2016.

\end{thebibliography}
\end{document}